# Classification of Diabetic Retinopathy via Fundus Photography: Utilization of Deep Learning Approaches to Speed up Disease Detection

Hangwei Zhuang
*Dept. of biomedical engineering*
Columbia University
New York, USA
hz2577@columbia.edu

Nabil Ettehadi
*Dept. of biomedical engineering*
Columbia University
New York, USA
ne2289@columbia.edu

*Abstract*—In this paper, we propose two distinct solutions to the problem of Diabetic Retinopathy (DR) classification. In the first approach, we introduce a shallow neural network architecture. This model performs well on classification of the most frequent classes while fails at classifying the less frequent ones. In the second approach, we use transfer learning to re-train the last modified layer of a very deep neural network to improve the generalization ability of the model to the less frequent classes. Our results demonstrate superior abilities of transfer learning in DR classification of less frequent classes compared to the shallow neural network.

*Keywords*—*Diabetic retinopathy, Transfer learning, Efficient net, Deep learning, Classification, Medical image processing*

## I. Introduction

In 2019, approximately 463 million adults (20-79 years) were living with diabetes and by 2045, this number is predicted to rise to 700 million [1]. According to the World Health Organization (WHO), the most common type of diabetes is type 2 diabetes, which occurs when the body becomes resistant to insulin or does not produce enough insulin [2]. Diabetic Retinopathy (DR) is a serious sight-threatening complication of diabetes, which affects blood vessels in the retina [3]. Nearly all type 1 diabetic patients develop retinopathy during the two decades of disease. Up to 21% of patients with type 2 diabetes have retinopathy at the time of their first diagnosis of diabetes, and most develop some degree of retinopathy over time. DR is the most frequent cause of new cases of blindness among adults aged 20–74 years [4]. Depending on the presence of clinical features, DR is classified into four types, namely mild Non-Proliferative Diabetic Retinopathy (NPDR), moderate NPDR, severe NPDR, and Proliferative DR (PDR) [5].

The primary method for evaluating DR involves direct and indirect ophthalmoscopy [6], [7]. Retinal imaging techniques such as fundus photography [8], optical coherence tomography [9], and fluorescein angiography [10] are utilized to assess the severity of DR in ophthalmoscopy [6]. These methods are used by highly trained specialists to diagnose and assess the severity of DR via manually reviewing the eye images. Such assessments are costly, time consuming, not easily accessible for every patient, and suffer from subjective opinions variance. To reduce the time and cost of the screening, automated algorithms are developed to analyze the images acquired. Neural networks [11]-[14], k-NN classifiers [15], and SVM [15], [16], along with various feature enhancing methods such as matched filtering, region growing, thresholding or optimal wavelet transform, are used for different classification problems of DR severity [5]. Consequently, high accuracies have been achieved in problems of two-class (normal vs DR), three-class (normal, NPDR and PDR), and four-class (normal, moderate NPDR, severe NPDR and PDR) classification using these methods. To the best of our knowledge, five-class (normal, mild, moderate, severe, and PDR) DR classification has sub-optimal results, with reported average accuracy of 85%, sensitivity of 82% and specificity of 86% [17].

Asia Pacific Tele-Ophthalmology Society (APTOS) 2019 Blindness Detection is a competition on Kaggle seeking machine-learning methods to automate and speed up DR detection [18]. With the hope of making the diagnosis of DR more accessible in rural areas, 2931 teams across the globe participated in this competition using a dataset of retina images. The retina images were taken by technicians from Aravind Eye Hospital in India, and manually labeled with DR severity via highly trained eye specialists.

Motivated by this competition, in this paper, we propose two deep-learning-based methods to address the classification of 5 class DR problem. In our first approach, we develop a novel shallow neural network architecture, while in the second approach we adopt a known architecture for transfer learning [19]. Our results demonstrate different abilities for the two the networks in terms of distinguishing different classes of DR.

## II. Methods and Data

### A. Data and Pre-processing

The dataset consists of 3662 colored RGB fundus photography images in PNG format, available publicly on Kaggle [18]. All images are labeled with five classes according to DR severity: 0 for no DR, 1 for mild, 2 for moderate, 3 for severe, and 4 for proliferative DR. The vast majority of the images have no or moderate DR. As illustrated in Table I, the dataset is highly unbalanced with 1805 no DR, 370 mild, 999 moderate, 193 severe, and 295 proliferative DR images. We split the whole dataset into train, validation, and test sets using a 0.8-0.1-0.1 split-ratio.

The images are taken under various lighting conditions, with different imaging devices, and from multiple clinics. As a result, images in the dataset vary in size, brightness and occasionally, focus. Some sample images are shown in Fig. 1 to demonstrate the variety of the raw images. To address the inconsistencies and to enhance relevant features, we used preprocessing steps inspired by [20]. First, we cropped out the black bands on the edges so that the field of view filled the whole image, reducing potential effects of irrelevant information. Next, all images were resized to 512×512, enabling size consistency of inputs to our models. Using a Gaussian kernel with sigma=10, the local mean color was subtracted from each image to increase local contrast and achieve a good color constancy [21]. Sample results of the preprocessing step are shown in Fig. 1 in contrast to the raw images.

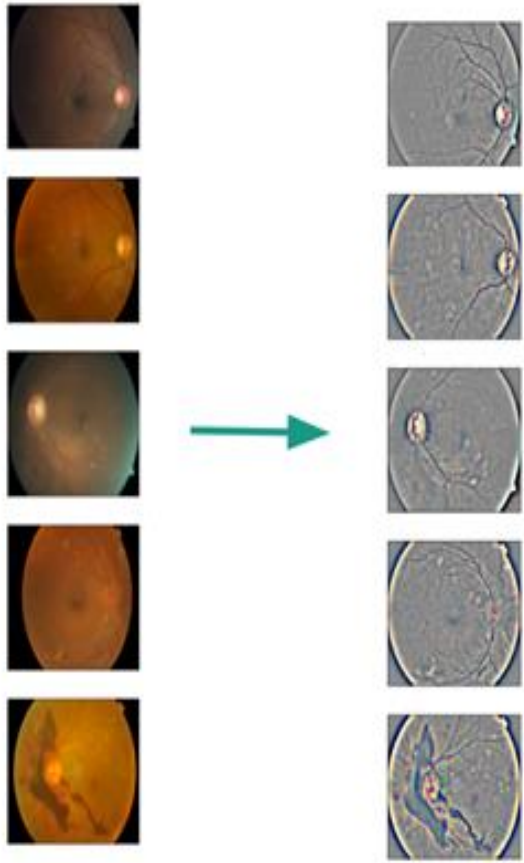

Fig. 1. Sample images before (left) and after (right) preprocessing inspired by [20]

TABLE I. NUMBER OF IMAGES IN EACH CLASS

| Class Label | Number of Images | Percentage of the class in the dataset |
|---|---|---|
| 0 | 1805 | 49.29% |
| 1 | 370 | 10.10% |
| 2 | 999 | 27.28% |
| 3 | 193 | 5.27% |
| 4 | 295 | 8.06% |

*B. Proposed Methods*

As previously discussed in the introduction section, we proposed two methods for solving this problem. In this section, the details of each method and the architectures of their underlying neural networks are discussed in details.

*B.1. Method 1: A shallow neural network*

In our first attempt at addressing this problem, we designed and trained a novel, relatively shallow neural network with 9 blocky-layers as depicted in Fig. 2. The network takes a batch of RGB (i.e., 3 channels) images of the size 512×512 as the input, and feeds them through multiple repeated blocky-layers of two types: I) a 2D convolutional layer, followed by a batch normalization layer, a RELU activation layer, and lastly a drop out layer, and II) a 2D Max-pooling layer, followed by a drop out layer. Finally, the tensors go through an unrolling (flatten) layer, followed by 2 fully connected (dense) layers that produce the outcome of the classifier (i.e., the class labels) via a SoftMax unit. Weights of all convolutional kernels and the two fully connect layers were initialized using the Glorot normal initializer [22]. Convolutional kernels for the 3 blocky-layers of type I were of sizes 13×13, 11×11, and 7×7 in the order of appearance, with valid padding (*i.e.,* no padding). Max-pooling kernel size of 2×2 was used for all 3 blocky-layers of type II. We used L2 regularizations for both the convolutional and fully connected layers, along with drop out layers to prevent overfitting to the training set [23], [24]. All of the activation functions used in the network were RELU except for the last fully connected layer, for which we used a SoftMax unit to get the class probabilities. Our model has a total of 26,808,133 parameters of which 192 were non-trainable. We used a weighted categorical cross-entropy loss function with more relative weights on the less frequent classes. For optimization purposes, we chose Adam optimizer [25] with the initial learning rate of 0.0001, and the default values introduced in [25] for the rest of the parameters. The batch size used for the training was set to 32.

*B.2. Method 2: Transfer learning*

A transfer learning approach was taken in our second attempt to solve the classification problem. We utilized Efficientnet-B3 designed in [26] and implemented in Pytorch by [27]. The Efficientnet-B3 network was pre-trained on ImageNet competition [28] and achieved 81.7% accuracy with fewer parameters than its competitors. Efficientnet-B3 is scaled up using compound scaling described in [26] from a baseline model consisting of MBConv blocks [29], and fully connected layers. Each MBConv block has a convolutional layer that expands the channels, a depth-wise convolutional layer with kernel size 3×3 or 5×5, and a convolutional layer that squeezes the channels. Each convolutional layer is followed by a batch normalization and is activated with memory-efficient swish activation function [30]. We adopted Efficientnet-B3 and modified its last layer according to the number of DR classes (i.e., 5). The modified architecture of Efficientnet-B3 is illustrated in Fig. 3., in which the last MBConv block is followed by a 2D adaptive average pooling layer, a dropout layer, and a linear layer to map the features into the 5 classes. In our model, we froze all layers except for the very last linear layer which was adapted and trained to output 5 classes of DR (instead of 1000 classes corresponding to the ImageNet competition). The final model has 10,703,917 parameters, of which 7685 are trained (the rest are frozen), allowing a better performance under limited computational resource. Similar to method 1, we used a weighted categorical cross entropy loss function with more weights on the less frequent classes. Unlike method 1, in this method we used a Stochastic Gradient Decent (SGD) optimizer with momentum. An initial learning rate of 0.01 along with a momentum coefficient of 0.9 was used during the optimization. The learning rate reduced on plateau at factor 0.85 and patience 2. The batch size was set to 64.

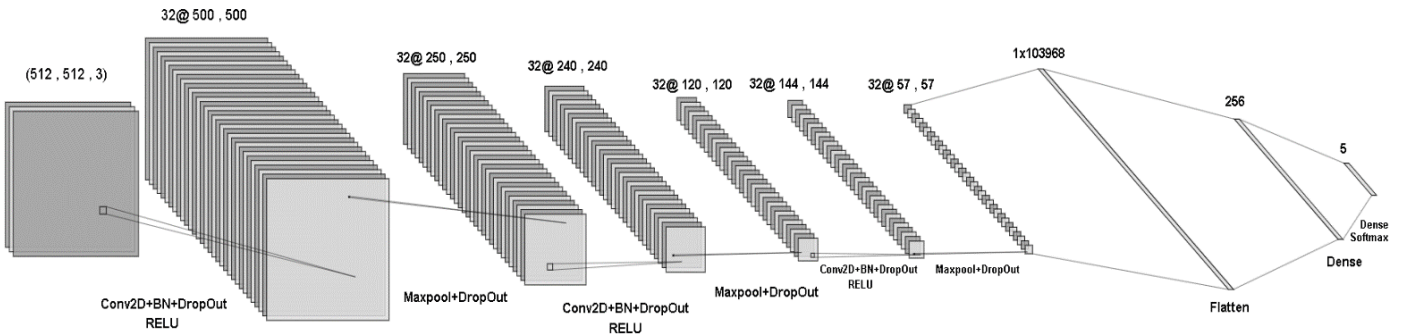

Fig. 2. Architecture of the neural network proposed in Method 1

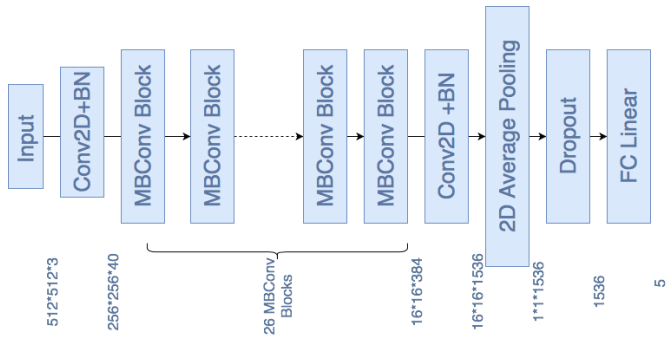

Fig. 3. Architecture of the modified Efficientnet-B3. Output tensor sizes are notated below blocks. All activation functions are memory-efficient swish.

## III. RESULATS AND DISCUSSION

### A. Experimental Results

#### A.1. Method 1:

We trained the network for 200 epochs on a Tesla K80 Google Cloud Platform GPU. The class weights for the loss function were set to [class 0: 1.2, class 1: 6.2, class 2: 3, class 3: 12.5, class 4: 8.2] via fine-tuning the model. As depicted in Fig. 4 the train loss monotonically decreased over epochs, whereas the validation loss showed a trend of decreasing, reaching a minimum, and increasing afterwards. Fig. 5 shows the classification accuracies of both train and validation sets. Both accuracies showed an overall increase followed by a saturation to their final steady values. The best model was chosen as the one that provides the minimum loss on the validation set. The metric used to measure the best model's performance was its accuracy in terms of classifying the classes. The model's accuracies on the train, validation, and test sets were 72.35%, 67.05%, and 69.03%, respectively. Fig. 6 shows the confusion matrix for the best model. As depicted in Fig. 6, the model performs well on class 0 and 2 with prediction accuracies of 92% and 77%, respectively. However, the model overfits to the two most frequent classes (i.e., 0 and 2), as the other classes (i.e., 1, 3, and 4) are mostly miss-labeled as class 2. This yields poor prediction accuracies of 15%, 21%, and 3% for class 1, 3, and 4, respectively.

#### A.2. Method 2:

The model was trained for 200 epochs on a Tesla K80 Google Cloud Platform GPU. The class weights for the loss function were empirically set to [class 0: 1, class 1: 3, class 2: 3, class 3: 5, class 4: 5] to achieve optimal results. As shown in Fig. 7, the train and validation loss decreased over time and reached their steady state values. As depicted in Fig. 8, both train and validation accuracies increased over epochs albeit small fluctuations. The best model was selected according to the highest validation accuracy. The best model has an accuracy of 80.85% on the train set, 80.60% on the validation set, and 77.87% on the test set. Fig. 9 shows the confusion matrix of the model's prediction. Overall, the matrix shows a better diagonal behavior, indicating relatively good accuracy and acceptable diagnostic power of the model. 97% of images with no DR were classified correctly and few images with DR were misclassified. Good predictions were made for class 2 (moderate DR), in which 73% are predicted correctly. However, accuracy on class 1, 3 and 4 are far from desired as they are less frequent in the train set. Only 41%, 30% and 56% accuracy were achieved for class 1,3 and 4 respectively, where a large number of images were misclassified as class 2.

### B. Discussion

We proposed two methods for the DR classification problem and detailed them in previous sections. In this section, we compare the results of the two methods. As depicted in Fig. 6 and Fig. 9, both methods show great accuracies for classification of class 0 (no DR) and class 2 (moderate DR). Although class 0 is the most frequent class in the training set, only few images in class 1 (2-3%), 2 (1-3%), and 3 (2%) are misclassified into class 0 for both methods. This indicates that both models perform excellently in a binary DR classification scenario (*i.e.*, DR vs no DR). Looking at the other elements of the confusion matrices in Fig. 6 and Fig. 9, it can be seen that both models have rooms for improvement in determination of the severity level of DR. However, it is evident that method 2 performs significantly better at predicting the labels of the less frequent classes (*i.e.*, class 1, 3, and 4) compared to method 1. Multiple reasons play roles for this significantly better performance: 1) Transfer learning takes advantage of previous training sessions. Efficientnet-B3 has been trained for very long hours over a much bigger training set (than the DR dataset). Hence, it has more power in image feature detection compared to the proposed model in method 1. Consequently, it can detect minor differences between different classes of DR more easily, resulting in a better classification accuracy for class 1, 3, and 4, compared to method 1. 2) Efficientnet-B3 is a much larger (*i.e.*, deeper) network compared to the shallow 9 blocky-layers network developed in method 1. Hence, method 2 has more power to learn more complicated features, mappings, and functions for classification purposes. The 1st place solution on Kaggle proposed a model using transfer learning and averaging 4 deep neural networks retrained on this dataset resulting in 85% accuracy on the test set [31]. Our proposed simple solution in method 2 has only about 7% less accuracy than the 1st place solution.

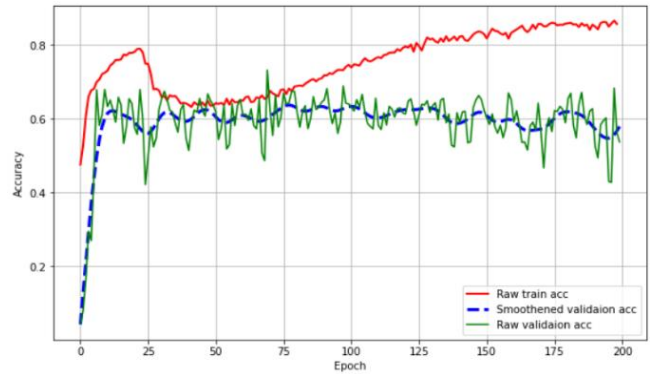

Fig. 5. Accuracy of the model in method 1 on the train and validation set over 200 epochs

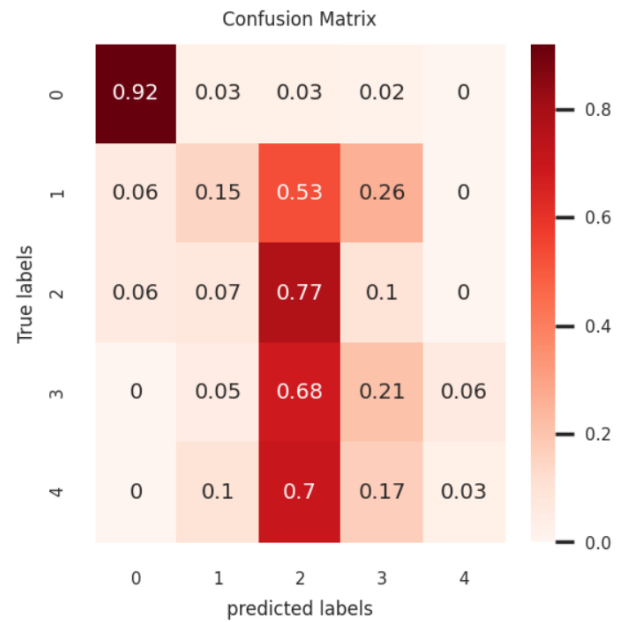

Fig. 6. Confusion matrix of the model in method 1

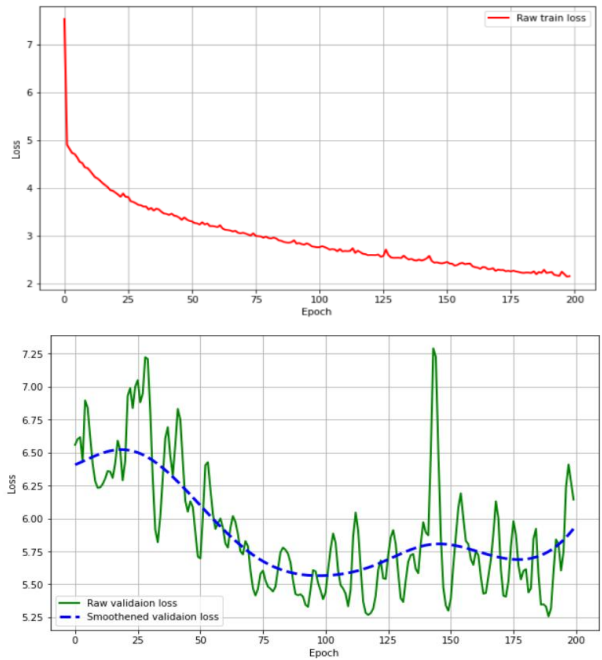

Fig. 4. Train (top) and validation (bottom) accuracies of the model in method 1. A smoother version of the raw validation loss is also depicted using a moving average window technique for a better visualization of the loss trend over 200 epochs.

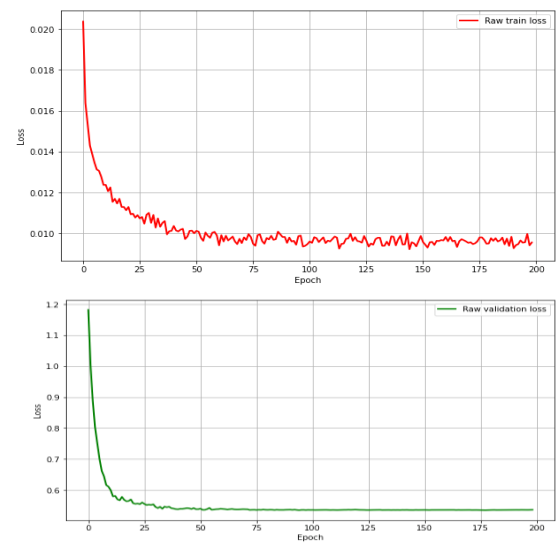

Fig. 7. Train (top) and validation (bottom) accuracies of the model in method 2 over 200 epochs

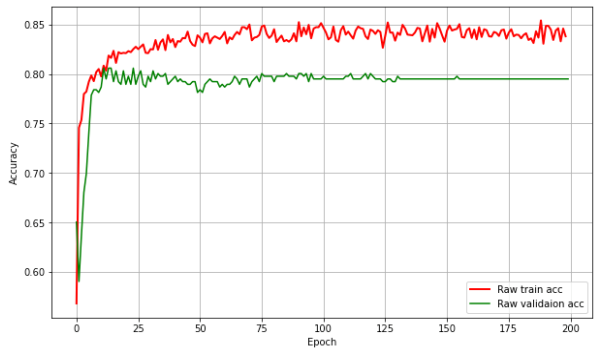

Fig. 8. Accuracy of the model in method 2 on the train and validation set over 200 epochs

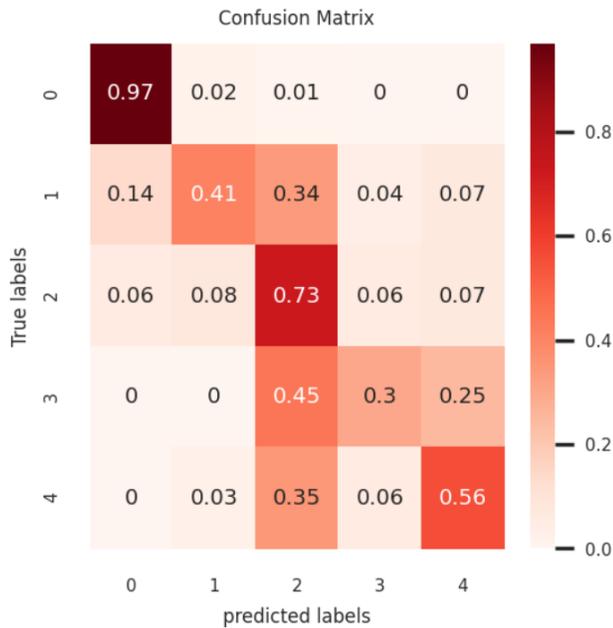

Fig. 9. Confusion matrix of the model in method 2

IV. CONCLUSIONS

In this paper, we developed two solutions to the DR classification problem. First, we proposed a novel shallow neural network architecture. This model performed well on the most frequent classes (i.e., 0 and 2). However, it could not generalize to the least frequent classes (i.e., 1, 3, and 4). In the second approach, we used transfer learning to re-train the last layer of Efficientnet-B3 (a much deeper network than the first solution) modified according to our classification data (*i.e.,* 5 categories instead of 1000). This model performed better compared to the first solution, as the classification accuracies of the least frequent classes increased significantly.

ACKNOWLEDGMENT

This project was inspired by APTOS 2019 Blindness Detection Kaggle competition [18] and was submitted as the final project for the course "BMEN E4460, Deep Learning in Biomedical Imaging", instructed by Dr. Andrew F. Laine at Columbia University. We thank him, Dr. Jia Guo, and all the other teaching assistant of this course for their great help.